# Simulation of optical flow and fuzzy based obstacle avoidance system for mobile robots

G.D. Illeperuma and D.U.J. Sonnadara

*Abstract*— Honey bees use optical flow to avoid obstacles effectively. In this research work similar methodology was tested on a simulated mobile robot. Simulation framework was based on VRML and Simulink in a 3D world. Optical flow vectors were calculated from a video scene captured by a virtual camera which was used as inputs to a fuzzy logic controller. Fuzzy logic controller decided the locomotion of the robot. Different fuzzy logic rules were evaluated. The robot was able to navigate through complex static and dynamic environments effectively, avoiding obstacles on its path.

## I. Introduction

Obstacle avoidance is a very important skill for a mobile robot. In this research, feasibility of using optical flow and fuzzy logic controller to avoid obstacles was tested. Most modern robots use an array of sensors such as LIDAR (light detection and ranging), IR or GPS to locate its position and avoid obstacles in its path. However above approach requires identifying the obstacles, its relative location and then to build an internal model of the world. This internal model is used to calculate a possible path for the robot [1]. This approach requires precise sensors and high computing power. In this research, a bio-inspired design is proposed based on the behavior of the honey bee. A honey bee can effectively avoid obstacles even with the low resolution monocular vision and few thousand neurons in the brain. It was discovered honey bees use optical flow to achieve this [2].

At present there are several optical flow based robots under development [3][4][5]. However the method proposed in this research is mostly based on subsumption architecture [6] not the classical internal model building approach. Hence it is expected to be fast and cost effective [7]. Inspired by the biological behavior, instead of using a PID controller, a fuzzy controller was used. The fuzzy controller allowed rapid testing of different control logics easily. A simulation framework was developed to test vision based robots and it was further improved to allow simulating and testing many types dynamic and static environments as well as robots. The virtual world was based on Virtual Reality Modeling Language (VRML) and hence supports many image properties such as texture and transparency. Finally, a physical robot was also build based on the controller algorithm which is yet be tested on the applicability under real world scenarios.

G.D. Illeperuma.
The Open University of Sri Lanka, Sri Lanka.

D.U.J. Sonnadara.
University of Colombo, Sri Lanka.

## II. Methodology

To compare different controller algorithms for the robot, it is necessary to benchmark all algorithms under identical conditions. It is also desirable to test these algorithms under different surroundings such as open spaces, cluttered offices and partially illuminated rooms. Therefore it was decided to use a simulation framework at the initial stage. The added advantage of the computer simulation is the ability to pause, rewind and debug the algorithm at will. It was also decided to simulate the complete process from image acquisition to locomotion control. Therefore, a framework that can simulate an interactive 3D world with image capturing capabilities was required. In addition, interfacing the 3D world with the controller algorithm was also required.

The simulation framework was able to simulate real world in 3D, with first persons or third person's perspectives. It also supports extensive visual details of objects such as the texture, amount of light reflected and transparency. Framework also supports easy integration with the MATLAB, which is used to implement the controller algorithm of the robot.

Structure of the simulation framework is described in Fig. 1.

### A. Testing Environment

The testing environment was constructed using VRML. There are three commonly used VRML versions. VRML, VRML 2.0 (VRML 97) and X3D [8]. Version used in this research was VRML 97 due to the compatibility with Simulink and it follows International Standard ISO/IEC 14772-1:1997 standard.

Different environments were created as VRML worlds such as corridors, mazes and open grounds. A small 3D model of the robot was created to visually track the behavior of the robot.

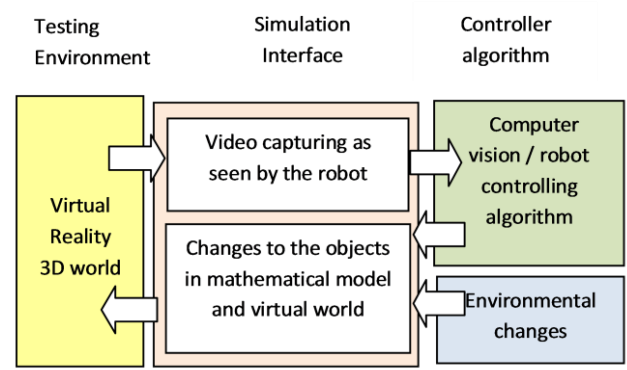

Figure 1.   Structure of the simulation framework

### B. Simulation interface sub system

Simulation interface sub system converted high level command signals to low level controller signals. It also performed the coordinate transformations. The video stream from the VRML was formatted to suit the controller algorithm.

### C. Environmental changes sub system

This module was responsible for the dynamic behavior of other objects in the world except the robot. For example it can simulate a car coming toward the robot. It was programmed separately and robot controller algorithm had no direct knowledge of the behavior of other objects.

### D. Controlling algorithm sub system

The function of the controller algorithm sub system was to capture the video stream as seen by the robot, process it, decide the next steps and send the control signal to the interface sub system. It consisted of image preprocessing, optical flow calculation, fuzzy controller and a post processing modules. The controller algorithm was implemented using Simulink and MATLAB.

### E. Image processing and optical flow calculation

The virtual camera captures images with $200 \times 320$ pixel resolution and color video stream with 10 frames per second rate. Image preprocessing step included converting the image into grayscale and applying image enhancements, such as contrast adjustments and histogram equalization. The processed image was passed to the optical flow calculating block.

Optical flow is the amount of image motion perceived by a human, an animal or a machine. There are several methods to calculate optical flow including phase correlation methods, Fourier transformation methods and differential methods [9]. Lucas–Kanade algorithm assumes, neighboring points of a local region centered around a point typically belong to the same surface hence make the same motion. Horn–Schunck algorithm assumes that the brightness of any part of the image world changes very slowly, so that the total derivative of the brightness is zero [10]. In this research both Lucas–Kanade algorithm and Horn–Schunck algorithm were tested.

The calculated optical flow values were segmented to three regions of interest (see Fig. 2); Left (L), Middle (M) and Right (R). Magnitude of each optical flow vector was calculated and for each region, three largest and three smallest values were removed. The concept behind this is to reject the extremely large and extremely small values created by the noise. From the remaining values, mean magnitude was calculated for each region and they were used as the inputs of the fuzzy controller.

### F. Fuzzy controller

Most decisions taken by humans or animals are not based on precise values but on fuzzy values. The behavior of bees to avoid obstacles is better described as "slow down if the speed is high" rather than an exact mathematical expression [11]. Such controllers are easily implemented as a fuzzy controller rather than a conventional program.

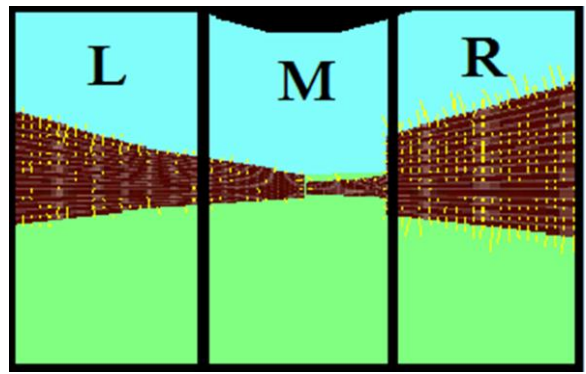

Figure 2.   Calculated optical flow vectors

Therefore a Fuzzy controller was selected and implemented using MATLAB fuzzy control toolbox.

The optical flow values were scaled to have a universe of discourse in the region of 0 - 10. Two models of fuzzy controller were modeled and tested. All the membership functions are based on 'smf' and 'zmf' membership functions [12].

Three controlling algorithms were tested for the performance. The first algorithm was based on 'center flying' concept. Here, the robot tries to balance the left side and right side optical motion. The input to the model, 'L_minus_R', was the difference of the optical flow on the sides. If the flow of one side is higher, robot would move to the opposite side by changing its output variable 'angle'.

Fuzzy rule set was defined as

1. If L_minus_R is R_Close then angle is 'turn_left'.

2. If L_minus_R is L_Close then angle is 'turn_right'.

Second algorithm was based on an improvement of the first algorithm allowing the robot to control the forward speed based on the total optical flow observed. Robot would slow down if the total optical motion is high. In the following, L_plus_R represent the total optical flow and speed is the output parameter.

3. If 'L_plus_R' is low_flow then speed is slow.

4. If 'L_plus_R' is high_flow then speed is fast.

The third model is called 'turn at threshold' and was used to overcome the 'tunnel syndrome' short coming of the 'center flying' model. It only change the direction if the optical flow of a side of the robot is too high, and slow down only if the both sides of the robot had very high optical flow values.

New input variables were left optical flow and right optical flow. Rule set was changed so that

1. If left flow is high and right flow is not high then angle is turn right.

2. If left flow is not high and right flow is high then angle is turn left.

3. If left flow is high and right flow is high then speed is slow.

4. If left flow is not high and right flow is not high then speed is fast.

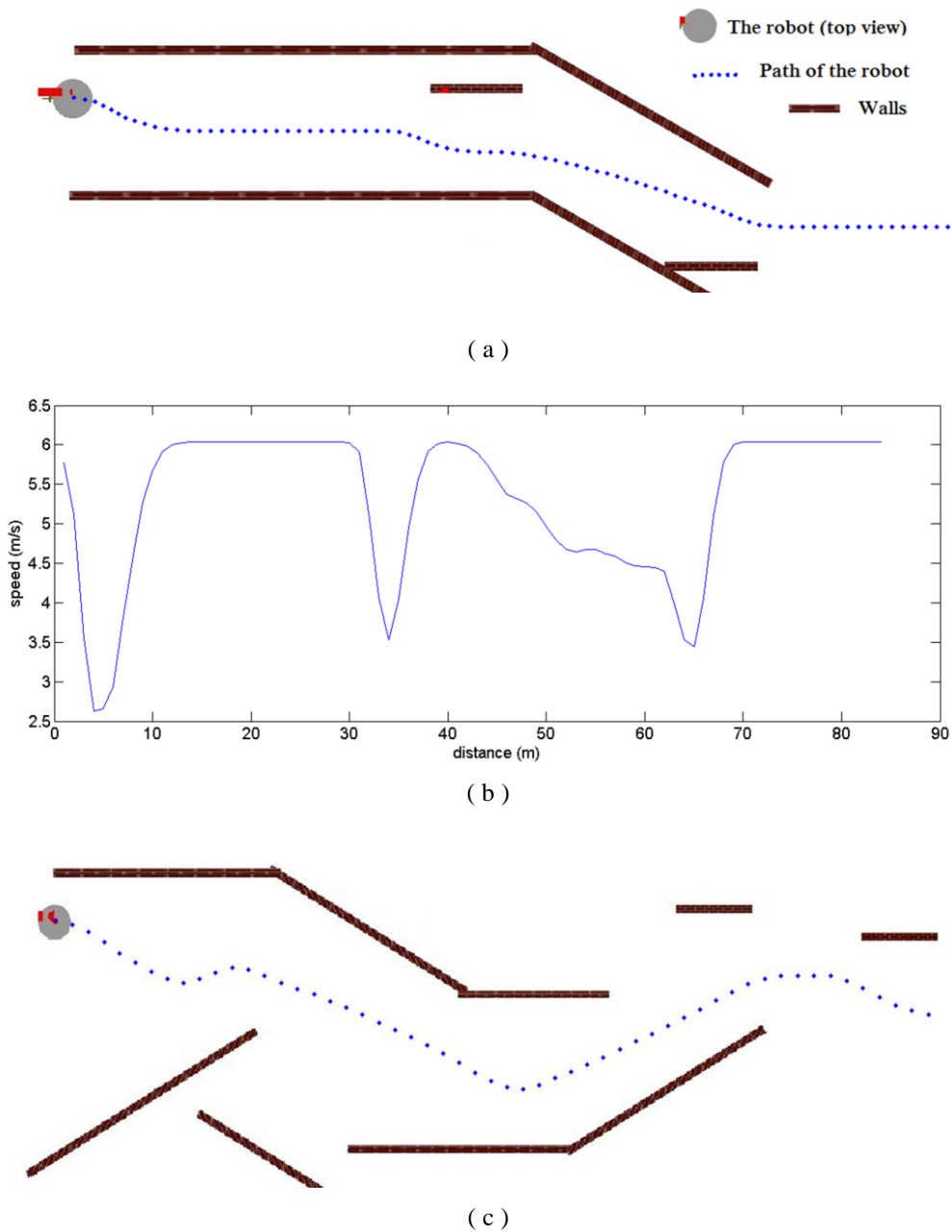

Figure 3.  (a) Path with center balanced model with speed control
(b)Speed variation inside the corridor (c) Robot following a complex path

### G. *Post processing*

Outputs of the fuzzy controller were de-normalized by multiplying by a weight factor. This allowed adjusting the sensitivity and impact without having to redefine the membership functions. At the post processing stage signals were also multiplexed to support VRML data structures.

## III. Results and Discussion

The first controller algorithm, 'center flying', was able to keep the robot at the center of a corridor. The algorithm was also tested with corridors with different sizes and angles and was found to be reasonably accurate.

However two draw backs were observed in this model. When the tunnel had a steep turn, robot did not change its direction fast enough to avoid it. It could be avoided by changing the weight factor. But it introduced the overshooting and exceeded physical constraints on the side velocity. A more logical approach was to slow down the forward motion of the robot when a corridor was narrow. This was achieved by adjusting the speed of the robot based on the total optical flow.

The second major drawback was its tendency to move toward open spaces, 'tunnel syndrome'. Since the robot was trying to balance side optical flows, robot looks for the maximum distance from obstacle on either side. This forced the robot to make unnecessary movement and also directed the robot into open holes in the corridor. To overcome this, the third model was introduced where robot only changed the direction if the side optical flow is high.  Fig. 3 shows the result of these models. It can be seen that the robot slows down when it is changing the direction or when the corridor is narrow. Comparatively, the robot makes more close turns and avoids unnecessary motions.

The same robot was tested in dynamic worlds where the walls and other obstacles were moving. The robot was able to evade them successfully.

Several draw backs were also identified. Uniform surfaces and dark walls prevented identification of features and thus reduced the optical flow values in the given region. Rotating objects also became a challenge since different parts of the same object produced different optical flow values. The technique discussed here needs to be modified if it is to be applied to a non-omni directional vehicle such as a micromouse. Instead of shifting to side these robots rotates which itself creates a horizontal velocity component on the optical flow. However by calculating the ego motion, adding latency to reaction time [5] or using curvature weighted depth problem of this extra optical flow can be solved [13].

Based on the captured data, at present, a Quadcopter is being automated to navigate avoiding obstacles.

## IV. Conclusions

In this research a fuzzy controller is integrated with optical flow vectors to design an autonomous obstacle avoidance system. Initial system was tested on a virtual environment. Several different designs including 'center following' and 'threshold avoiding' were tested and found to be successful at avoiding obstacles effectively. In general it is concluded that it is feasible to use this method in mobile robots., It is expected to compare the simulated results with a physical experiment results in the near future.

### *Acknowledgment*

Financial assistance by the University of Colombo Research Grant number 2012/CG/14 is acknowledged.

About Author (s):

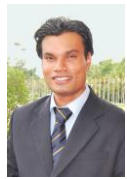

Gayan Illeperuma obtained his B.Sc. degree in Computational Physics (Hons) and B.IT degree from the University of Colombo, Sri Lanka. He is currently working as a Lecturer at the Department of Physics, The Open University of Sri Lanka and engaged in research relating to the application of computer vision in mobile robots.

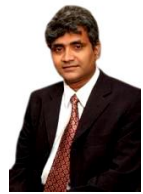

Upul Sonnadara obtained his B.Sc. degree in Physics (Hons) from the University of Colombo, Sri Lanka. He obtained his Ph.D. degree in Relativistic Heavy Ion Physics from the University of Pittsburgh, USA. At present he works as a Senior Professor at the Department of Physics, University of Colombo and engaged in research related to instrumentation and computational science.